\documentclass[letterpaper, 10 pt, conference]{ieeeconf}  

\IEEEoverridecommandlockouts                              

\usepackage{graphics} 
\usepackage{epsfig} 
\usepackage{mathptmx} 
\usepackage{times} 
\usepackage{amsmath} 
\usepackage{amssymb}  
\usepackage{hyperref}
\usepackage{subcaption}
\usepackage{caption}

\makeatletter
\newcommand*\titleheader[1]{\gdef\@titleheader{#1}}
\AtBeginDocument{%
  \let\st@red@title\@title%
  \def\@title{%
    \bgroup\normalfont\large\centering\@titleheader\par\egroup
    \vskip1.5em\st@red@title}
}

\title{\LARGE \bf
Individualised Treatment Effects Estimation with Composite Treatments and Composite Outcomes
}
\titleheader{The 47th Annual International Conference of the IEEE Engineering in Medicine and Biology Society (to appear).}

\author{Vinod Kumar {Chauhan}$^{1}$, Lei Clifton$^{1,2}$, Gaurav Nigam$^{1,3}$ and David A. Clifton$^{1,4}$
\thanks{$^{1}$Vinod Kumar {Chauhan} is with the Institute of Biomedical Engineering at the University of Oxford UK, and is the corresponding author: \href{mailto:vinod.kumar@eng.ox.ac.uk}{vinod.kumar@eng.ox.ac.uk}.
        }%
\thanks{$^{1,2}$Lei Clifton is with the Nuffield Department of Primary Care Health Sciences at the University of Oxford UK and is also affiliated to the Institute of Biomedical Engineering: \href{mailto:lei.clifton@phc.ox.ac.uk}{lei.clifton@phc.ox.ac.uk}.
        }%
\thanks{$^{1,3}$Gaurav Nigam is with the Nuffield Department of Medicine at the University of Oxford, UK and is also affiliated to the Institute of Biomedical Engineering: \href{mailto:gaurav.nigam@wadham.ox.ac.uk}{gaurav.nigam@wadham.ox.ac.uk}.}
\thanks{$^{1,4}$David A. Clifton is with the Institute of Biomedical Engineering at the University of Oxford UK and is also affiliated to the Oxford-Suzhou Institute of Advanced Research (OSCAR), Suzhou, China: \href{mailto:david.clifton@eng.ox.ac.uk}{david.clifton@eng.ox.ac.uk}.
        }%
}

\begin{document}

\maketitle
\thispagestyle{empty}
\pagestyle{empty}

\begin{abstract}
    Estimating individualised treatment effect (ITE) -- that is the causal effect of a set of variables (also called exposures, treatments, actions, policies, or interventions), referred to as \textit{composite treatments}, on a set of outcome variables of interest, referred to as \textit{composite outcomes}, for a unit from observational data -- remains a fundamental problem in causal inference with applications across disciplines, such as healthcare, economics, education, social science, marketing, and computer science. Previous work in causal machine learning for ITE estimation is limited to simple settings, like single treatments and single outcomes. This hinders their use in complex real-world scenarios; for example, consider studying the effect of different ICU interventions, such as beta-blockers and statins for a patient admitted for heart surgery, on different outcomes of interest such as atrial fibrillation and in-hospital mortality. The limited research into composite treatments and outcomes is primarily due to data scarcity for all treatments and outcomes.
    To address the above challenges, we propose a novel and innovative hypernetwork-based approach, called \emph{H-Learner}, to solve ITE estimation under composite treatments and composite outcomes, which tackles the data scarcity issue by dynamically sharing information across treatments and outcomes. Our empirical analysis with binary and arbitrary composite treatments and outcomes demonstrates the effectiveness of the proposed approach compared to existing methods.
    
    {\textbf{\textit{Clinical Relevance}}}\textemdash This paper develops a novel methodology, the H-Learner, to estimate individualised treatment effects when clinicians need to consider multiple interventions simultaneously to impact several patient outcomes. By accurately estimating these complex, individualised effects from observational data, this method has the potential to provide clinicians with more precise insights for tailoring treatment strategies, ultimately leading to improved patient care and outcomes in complex clinical scenarios where single interventions and outcomes are insufficient to capture the full picture.
\end{abstract}

\section{INTRODUCTION}
\label{sec_intro}
Causal inference helps to answer fundamental ``what if'' questions regarding the effect of a set of variables on a set of outcome variables of interest \cite{Chauhan2025beyond}. For instance, one might be interested in estimating the risk of cancer for an individual if they were to quit smoking. Such questions are prevalent in science, including healthcare, economics, education, social science, marketing, and computer science. While randomised controlled trials remain the gold standard for causal effect estimation, they can be time-consuming, expensive, ethically challenging, and more importantly not suitable for individual-level causal effect estimation due to their strict inclusion criteria and small and less representative samples. By contrast, non-experimental approaches based on observational datasets (e.g., electronic health records) address many of these limitations and can complement randomised controlled trials. Recently, there has been notable progress in causal machine learning \cite{bica2021real,chauhan2024dynamic}, a discipline that integrates rigorous causal inference methodologies with state-of-the-art machine learning techniques, particularly for the analysis of observational data.

\begin{figure}[htb!]
    \centering
    \includegraphics[width=\linewidth]{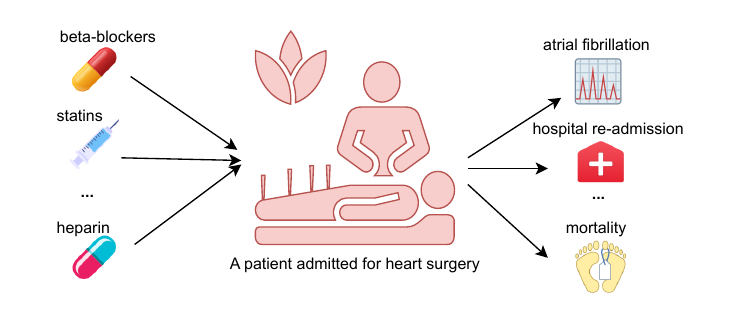}
    \caption{Example: An intensive care unit patient, admitted for heart surgery, needs composite treatments to optimise composite outcomes.}
    \label{fig_problem}
\end{figure}

Existing research in individualised treatment effects (ITE), also referred to as conditional average treatment effects estimation or heterogeneous treatment effects estimation in the machine learning literature \cite{shalit2017estimating}, is mostly focused on simple settings considering single treatments, such as binary treatments, and single outcomes \cite{curth2021inductive,chauhan2023adversarial,hernan2023causal,chauhan2024dynamic,melnychuk2025orthogonal}. However, the real world is complex and might involve \emph{composite treatments}, i.e., multiple simultaneous treatments (often referred to as polypharmacy in clinical literature in the context of medications \cite{hajjar2007polypharmacy}, or multimodal treatment more broadly), as well as \emph{composite outcomes}, where we aim to optimise multiple simultaneous outcomes of interest. Thus, existing methods are not always suitable for practical adoption in these multi-faceted scenarios.

\textbf{Motivating example}: Suppose a patient is admitted to a hospital for heart surgery. Approximately 20--50\% of such patients are at risk of atrial fibrillation within the first seven days after surgery \cite{d2018society}, which is associated with an increased risk of heart failure, dementia, and stroke \cite{Munger2014}. During and after the surgery, clinicians can choose from multiple interventions (e.g., beta-blockers, statins, DC cardioversion etc) to manage multiple outcomes, including atrial fibrillation, mortality, and readmission to the intensive care unit, as depicted in Fig.~\ref{fig_problem}. Evaluating the combined effects of these interventions on the outcomes of interest is a complex open problem that urgently requires solutions with potentially significant impact on healthcare.


The primary challenge with composite treatments and outcomes is the scarcity of data for certain treatment combinations. For instance, with $K$ binary treatments, there are $2^K$ possible treatment combinations, and in many real-world settings, it is impractical (or impossible) to obtain sufficient observations for all these combinations. This data deficiency hampers the reliable estimation of ITE for more complex treatment-outcome scenarios. Furthermore, analysing composite treatments and outcomes necessitates addressing confounding bias for each treatment component and outcome individually, compounding the challenge.

To address the above challenges in ITE estimation for complex real-world scenarios, we propose a novel and innovative approach based on a hypernetwork (a neural network that generates weights for another neural network) \cite{chauhan2024brief}, referred to as \emph{H-Learner}, to solve the problem of composite treatments and outcomes. In H-Learner, the hypernetwork is conditioned on the treatments and outcomes to generate a target learner for each specific treatment combination. Because H-Learner maps treatments to the corresponding target learners, it can tackle data scarcity by dynamically sharing information across treatments and outcomes. We provide an empirical study of H-Learner under varying treatments and outcomes, comparing its performance against existing methods.

\textbf{Contributions:} (i) We present an innovative and novel methodology, H-Learner, which leverages hypernetworks for ITE estimation under composite treatments and outcomes by sharing information across different treatment-outcome combinations; (ii) We empirically validate H-Learner against state-of-the-art approaches, showing superior or at par performance in scenarios involving varying treatments and outcomes with minimal assumptions.



\section{RELATED WORK}
\label{sec_related_work}
In this section, we briefly review the literature on ITE estimation and hypernetworks.

\subsection{Individualised Treatment Effects Estimation}
\label{subsec_ITE}
After the pioneering work of Johansson et al. \cite{johansson2016learning}, a wide variety of machine learning-based ITE learners have been proposed \cite{bica2021real}. These methods can be broadly categorised as: (i) representation learning or deep learning-based learners, e.g., \cite{curth2021inductive,shalit2017estimating,hassanpour2019counterfactual,hassanpour2019learning,chauhan2023adversarial,chauhan2024dynamic}, (ii) tree-based learners \cite{athey2019estimating}, and (iii) meta-learners, which are model-agnostic and can be subdivided into direct learners (S-Learner and T-Learner) \cite{kunzel2019metalearners}, and indirect learners (RA-Learner, DR-Learner, and X-Learner) \cite{kennedy2023towards,curth2021nonparametric,kunzel2019metalearners}. For a comprehensive review, please refer to Curth et al. \cite{Curth2024Machine}. While these frameworks have demonstrated success, they predominantly address settings with a single treatment and a single outcome.

The study of composite treatment effects has been constrained by data scarcity arising from the exponential number of treatment combinations (for $K$ binary treatments, $2^K$ combinations) and the possibility that many combinations have insufficient or no data at all. In this context, Wang et al. \cite{wang2019blessings} proposed an S-Learner-based approach, called deconfounder (DEC), that accommodates composite (binary and real) treatments for a single outcome. They leverage proxy variables and latent representations to capture interactions among multiple treatments and mitigate unmeasured confounding. Similarly, \cite{zou2020counterfactual} developed the variational sample re-weighting (VSR) approach—an extension of S-Learner—to handle high-dimensional binary treatments (referred to as bundle treatments), using learned latent representations to de-correlate treatments from confounders. To address data scarcity, \cite{qian2021estimating} proposed a data-augmentation strategy for composite treatment and single outcome settings, whereby $K$ ITE learners generate balanced datasets for each treatment before training a standard supervised deep learning model to predict potential outcomes. However, this method requires to train multiple models and is also limited to binary composite treatments.




\subsection{Hypernetworks}
\label{subsec_hypernetworks}
Hypernetworks, or hypernets, are a class of neural networks that generate the weights of another network, known as the target (or primary) network \cite{chauhan2024brief}. Although the core idea of context-dependent weight generation originated earlier \cite{schmidhuber1992learning}, the term ``hypernetwork'' was popularised by \cite{ha2017hypernetworks}, who introduced an end-to-end training paradigm for both the hypernetwork and the target network. Hypernetworks thus provide an alternative way to train neural networks \cite{chauhan2024brief}, which consist of a primary network for performing predictions and a hypernetwork for generating the primary network’s weights. Depending on the type of conditioning (data, task identifiers, or noise), hypernetworks are respectively categorised as data-conditioned, task-conditioned, or noise-conditioned hypernetworks. They can be classified using five key design criteria based on input-output variability and architectural choices \cite{chauhan2024brief}.

When a hypernetwork is used to generate weights for multiple target networks (referred to as soft-weight sharing), it enables end-to-end information sharing across those networks. Soft-weight sharing was recently used in \cite{chauhan2024dynamic} to propose HyperITE that addresses the problem of information sharing between potential outcome functions for training ITE learners. Nevertheless, HyperITEs are limited to binary treatments and single outcomes, unlike our approach which can handle an arbitrary number of treatments and outcomes.

Hypernetworks have emerged as a powerful deep learning technique due to their flexibility, expressivity, data-adaptivity, and information sharing, and have been used across various problems in deep learning \cite{chauhan2024brief}. For example, hypernetworks have been successfully applied and have shown better results across different deep learning problems, such as uncertainty quantification \cite{deutsch2019generative,Chan2024}, hyperparameter optimisation \cite{lorraine2018stochastic}, continual learning \cite{Oswald2020Continual}, federated learning \cite{shin2024effective}, multitasking \cite{tay2021hypergrid}, embedding representations \cite{yoo2024hyper}, ensemble learning \cite{kristiadi2019predictive}, multi-objective optimisation \cite{Tuan2024}, weight pruning \cite{liu2019metapruning}, model-extraction attack \cite{yuan2024hypertheft}, unlearning \cite{rangel2024learning}, image processing \cite{Ramanarayanan_2023_ICCV,Zhou2024}, quantum computing \cite{carrasquilla2023quantum}, knowledge distillation \cite{wu2023hyperinr}, neural architecture search \cite{peng2020cream}, adversarial defence \cite{sun2017hypernetworks}, and learning partial differential equations \cite{botteghi2025hyperl}. For an overview of hypernetworks, refer to \cite{chauhan2024brief}.


\section{BACKGROUND}
\label{sec_background}
\textbf{Notations:} Let $\mathbf{x}_i = (x_{i1}, \dots, x_{ip})^\top$ denote the $p$-dimensional vector of pre-treatment features for patient $i$ ($i=1, \dots, N$). Let $\mathbf{t}_i = (t_{i1}, \dots, t_{iK})^\top$ be the $K$-dimensional treatment vector for patient $i$, where $t_{ik}$ represents the $k$-th component of the potentially composite treatment. Let $\mathbf{y}_i = (y_{i1}, \dots, y_{iM})^\top$ be the $M$-dimensional outcome vector for patient $i$, where $y_{im}$ is the $m$-th outcome.  $y_{im}(\mathbf{t}_i)$ denotes the potential outcome for patient $i$ on outcome $m$ if they were to receive treatment vector $\mathbf{t}_i$.  The observed outcome is $y_{im} = y_{im}(\mathbf{t}_i)$. The observational dataset is $\mathcal{D} = \{(\mathbf{x}_i, \mathbf{t}_i, \mathbf{y}_i)\}_{i=1}^N$.

\textbf{Potential Outcomes Framework:} Following the potential outcomes framework \cite{rubin2005causal}, we define $y_{im}(\mathbf{t}_i)$ as the potential outcome for individual $i$ on outcome $m$ if they were to receive treatment vector $\mathbf{t}_i \in \mathcal{T}$.  This represents the counterfactual outcome that would have been observed had individual $i$ received treatment $\mathbf{t}_i$, even if they actually received a different treatment.  We can then represent the vector of potential outcomes under treatment $\mathbf{t_i}$ as $\mathbf{y}_i(\mathbf{t}_i) = (y_1(\mathbf{t}_i), y_2(\mathbf{t}_i), \dots, y_M(\mathbf{t}_i))^\top$.  The fundamental problem of causal inference is that we can only observe one potential outcome vector for each individual, namely $\mathbf{y}_i = \mathbf{y}_i(\mathbf{t}_i)$, where $\mathbf{t}_i$ is the treatment vector actually received by individual $i$.  Specifically, the observed outcome for individual $i$ on outcome $m$ is $y_{im} = y_m(\mathbf{t}_i)$.

\textbf{Assumptions:} To identify and estimate ITEs for composite treatments and composite outcomes, we rely on the standard causal assumptions of (i) unconfoundedness (or conditional exchangeability), (ii) positivity (or overlap) and (iii) consistency for all individual treatments and outcomes.

\section{METHODS}
\label{sec_methods}
In this section, we introduce \emph{H-Learner}, a hypernetwork-based approach to solve ITE estimation problems with composite treatment and composite outcome.


\textbf{Motivation:}
Complex real-world scenarios often entail multiple treatments and multiple outcomes of interest. Existing research on ITE estimation, however, typically focuses on single treatment and single outcome problems. In the composite treatment and composite outcome setting, the exponential number of possible treatment combinations (e.g., $2^K$ for $K$ binary treatments) leads to data scarcity, where certain treatment combinations may have limited or no samples. Such data sparsity poses a significant challenge to reliably estimating ITE.

Broadly, there are two ways to address composite treatment and composite outcome problems:  
\emph{(i) Train independent models for each treatment combination,} akin to a T-Learner \cite{kunzel2019metalearners}. While flexible, this approach can be infeasible when data are scarce for many treatment combinations.  
\emph{(ii) Train a single joint model for all treatment combinations and outcomes,} such as an S-Learner \cite{kunzel2019metalearners}. This approach can alleviate data scarcity through shared learning but may lack the flexibility needed even for simpler (binary) scenarios.  

To overcome these limitations, we utilise hypernetworks for joint training of ITE learners. Our proposed \emph{H-Learner} effectively combines the advantages of both independent and joint training: it dynamically shares information across treatments and outcomes yet learns distinct learners for each treatment-outcome combination, mitigating data scarcity.


\begin{figure}[htb!]
    \centering
    \includegraphics[width=0.65\linewidth]{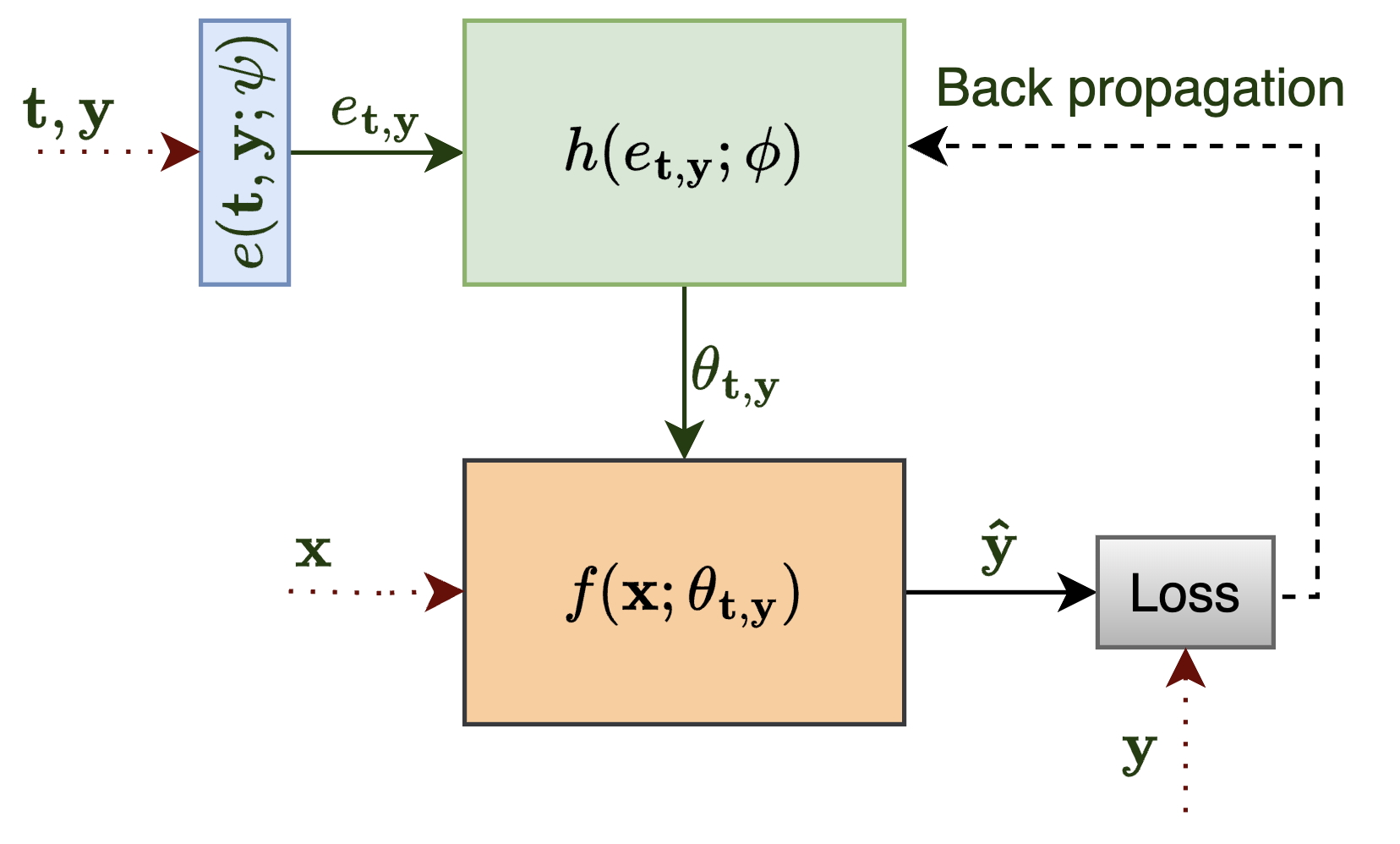}
    \caption{Architecture of H-Learner: It comprises an embedding layer $e(t,y;\psi)$, a hypernetwork $h(e_{\mathbf{t},\mathbf{y}}; \phi))$, and an ITE learner $f(\mathbf{X};\theta_{t,y})$. The hypernetwork $h$, conditioned on an embedding of treatment-outcome combination $e_{\mathbf{t},\mathbf{y}}$, generates the ITE learner’s $f$ weights $\theta_{t,y}$, enabling dynamic information sharing across treatments and outcomes.}
    \label{fig_HLearner}
\end{figure}

\begin{figure*}[htb!]
    \begin{subfigure}{0.32\textwidth} 
        \centering
        \includegraphics[width=\textwidth]{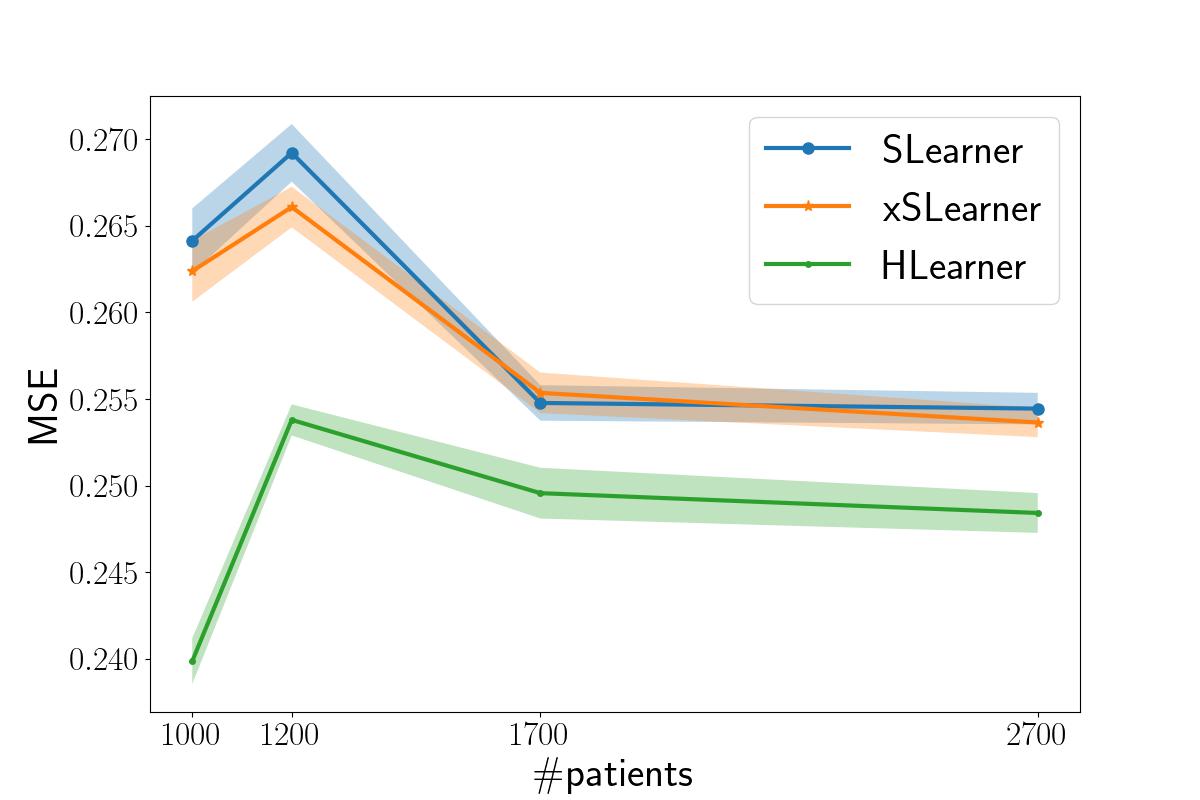} 
    \end{subfigure}
    \hfill 
    \begin{subfigure}{0.32\textwidth}
        \centering
        \includegraphics[width=\textwidth]{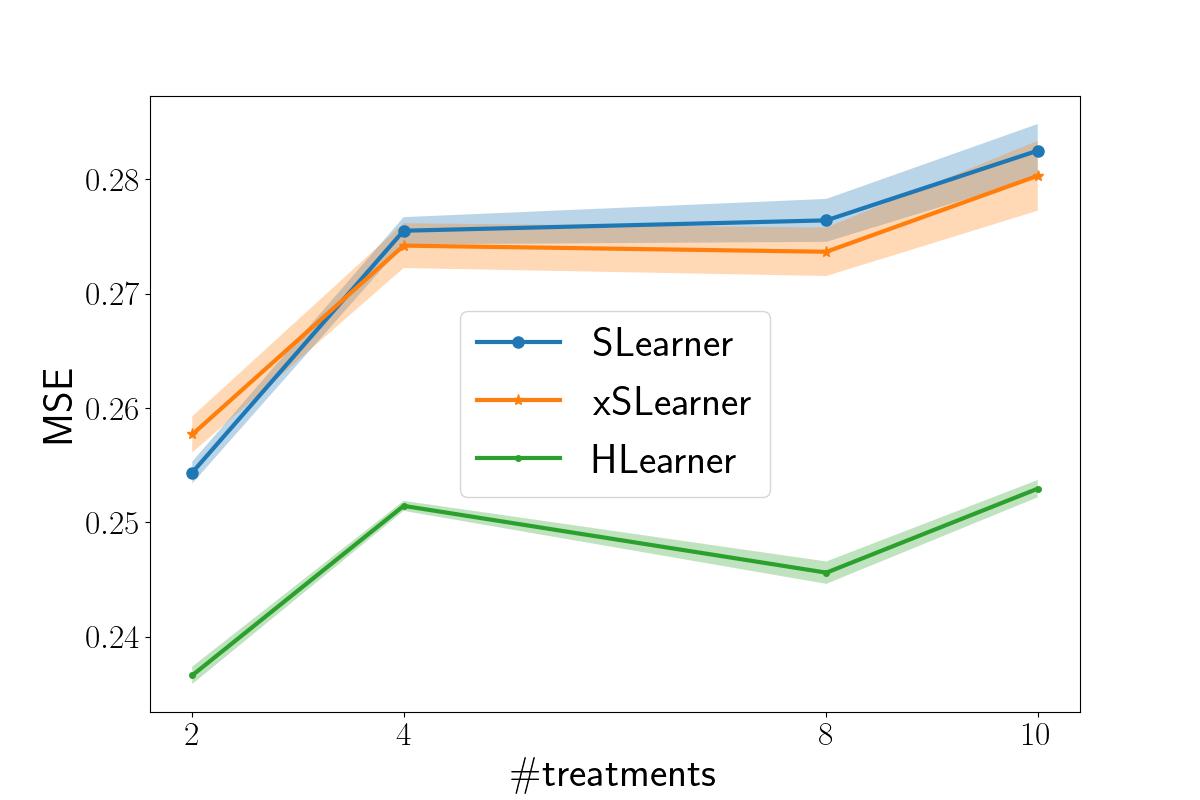}
    \end{subfigure}
    \hfill
    \begin{subfigure}{0.32\textwidth}
        \centering
        \includegraphics[width=\textwidth]{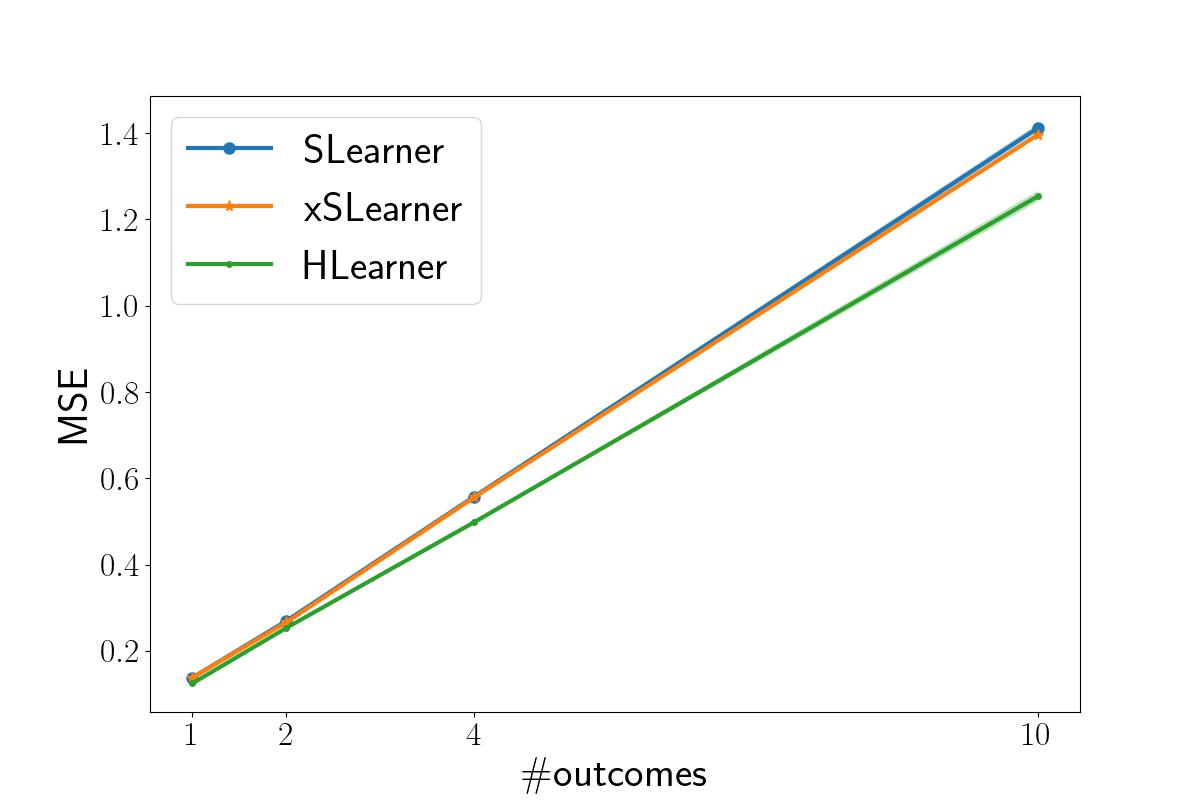}
    \end{subfigure}
    \caption{Comparative study of ITE learners for arbitrary composite treatments and composite outcomes (shaded region represents one standard error over 10 repetitions).}
    \label{fig_results_continuous}
\end{figure*}

\begin{figure*}[htb!]
    \begin{subfigure}{0.32\textwidth} 
        \centering
        \includegraphics[width=\textwidth]{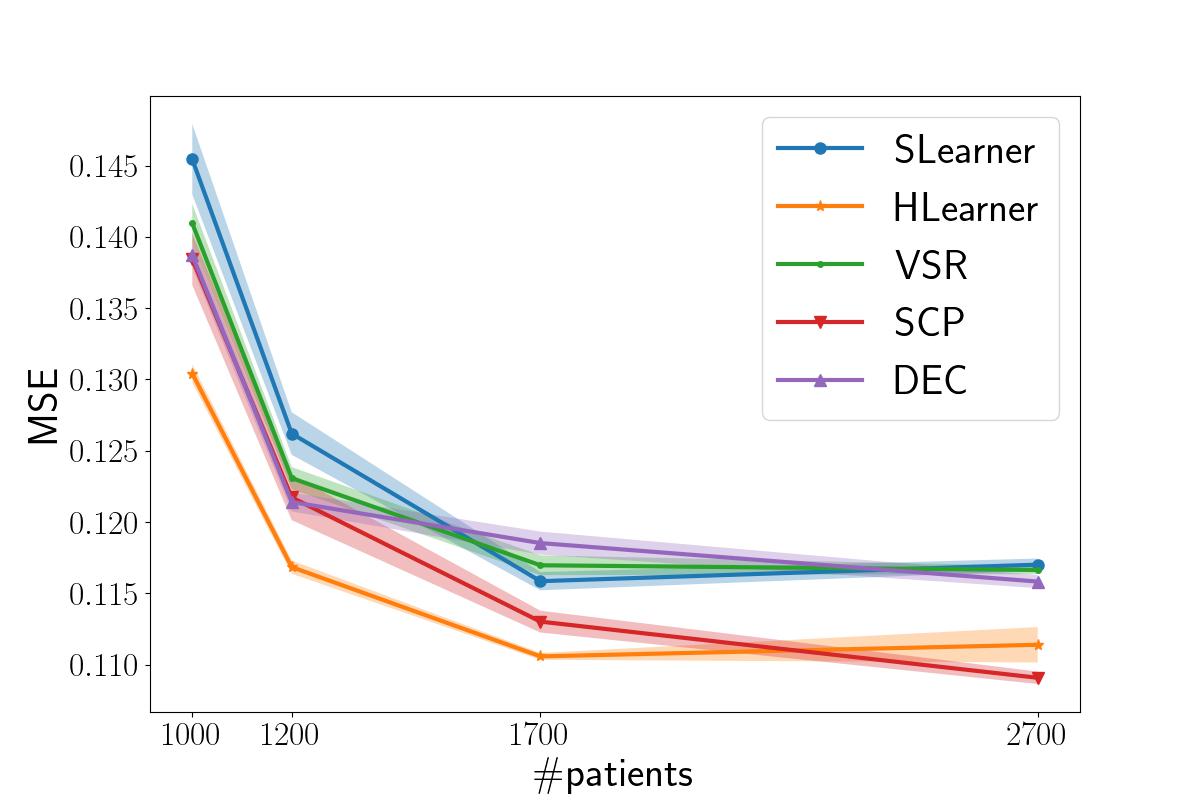} 
    \end{subfigure}
    \hfill 
    \begin{subfigure}{0.32\textwidth}
        \centering
        \includegraphics[width=\textwidth]{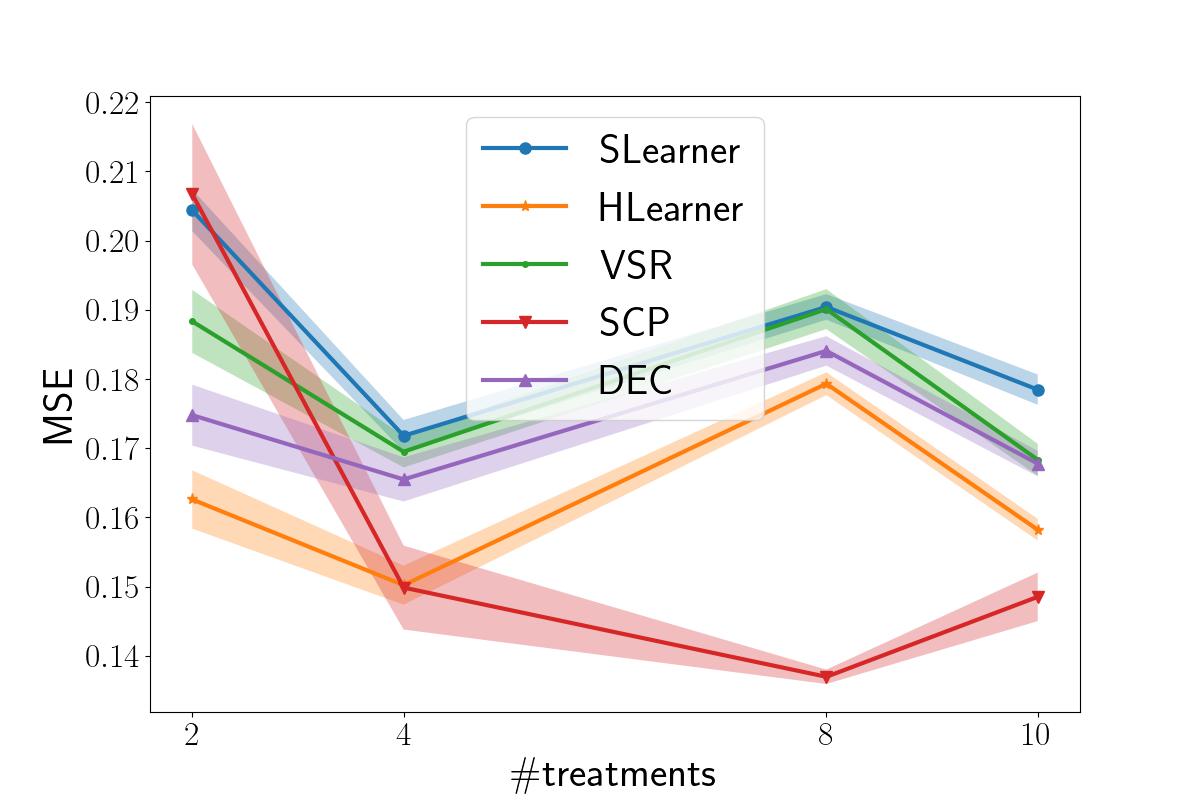}
    \end{subfigure}
    \hfill
    \begin{subfigure}{0.32\textwidth}
        \centering
        \includegraphics[width=\textwidth]{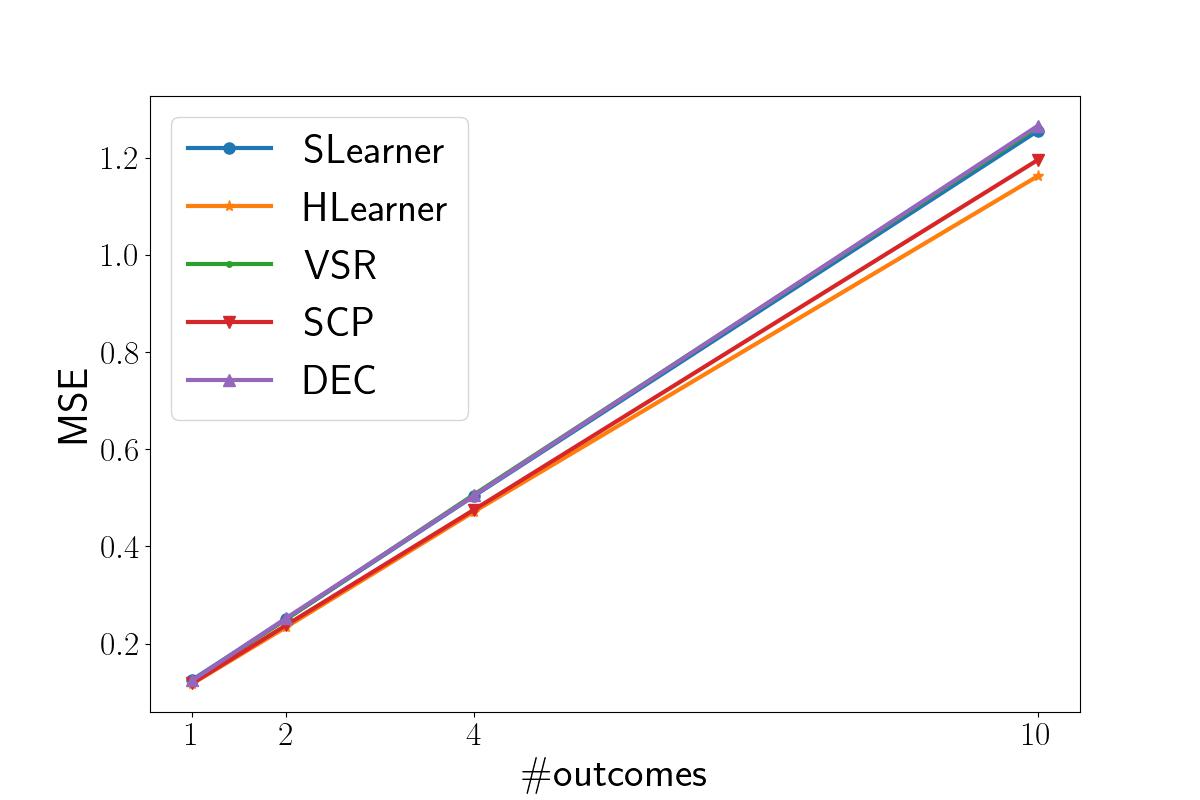}
    \end{subfigure}
    \caption{Comparative study of ITE learners for binary composite treatments and composite outcomes (shaded region represents one standard error over 10 repetitions). H-Learner does not need any assumptions about the interaction among the treatments and can handle composite outcomes, but SCP needs to know the causal structure among treatments and is limited to single outcomes.}
    \label{fig_results_binary}
\end{figure*}

\textbf{Neural Architecture and Working:} 
Figure~\ref{fig_HLearner} illustrates the key components of H-Learner, which include an embedding layer $e(t,y;\psi)$, a hypernetwork $h(e_{\mathbf{t},\mathbf{y}}; \phi))$, and a target network (the ITE learner) $f(\mathbf{X};\theta_{t,y})$, where $\psi,\; \phi$ and $\theta$ are weights of embedding layer, hypernetwork and the ITE learner, respectively. The embedding layer learns representations for each unique combination of treatments and outcomes. It converts each outcome into a one-hot encoded vector and passes it through a linear layer to get the outcome embedding. To get treatment embedding, the treatment combination is passed through a linear layer, and the final embedding for the outcome and treatment combination is obtained by concatenating the respective embeddings. The embedding layer maps given causes and an effect to a meaningful representation, which is equivalent to learning embeddings for words in natural language processing \cite{wang2020survey}.

The hypernetwork is a multi-layer perceptron that takes the treatment-outcome embeddings as input (i.e., \emph{task-conditioned} inputs \cite{chauhan2024brief}) and outputs the weights of the target network. The target network, in turn, receives the covariate data and predicts the potential outcomes for the specified treatment-outcome combination. Unlike conventional learners where the target network’s weights are trained directly, here they are entirely generated by the hypernetwork and thus are not themselves learned via backpropagation.

Training proceeds by feeding the treatments and outcomes into the embedding layer, whose outputs are passed to the hypernetwork. The hypernetwork produces weights for the target network, which processes the observed covariates to predict potential outcomes. We compute the prediction loss on factual observations (the data points for which the treatments and outcomes are known) and backpropagate the error. Since only the embedding layer and the hypernetwork parameters are trainable, the backpropagated gradients update those components to learn better embeddings for treatments and outcomes and improve the hypernetwork’s mapping from embeddings to target-network weights. Each training epoch thus leads to dynamic information sharing across the various treatments and outcomes, benefiting low-data configurations through mechanisms similar to transfer learning and multitasking \cite{maurer2016benefit}.


Formally, we define the optimisation objective for H-Learner as:
\begin{equation}
    \label{eq_hlearner}
    \min_{\phi, \psi} f(\mathbf{x}; \theta_{\mathbf{t},\mathbf{y}} = h(e_{\mathbf{t},\mathbf{y}}=e(\mathbf{t},\mathbf{y};\psi); \phi)), \quad \forall \{\mathbf{x}, \mathbf{t},\mathbf{y}\} \in \mathcal{D},
\end{equation}
where $\phi$ and $\psi$ are the trainable parameters of the hypernetwork and the embedding layer, respectively, $e(\cdot,\cdot;\psi)$ denotes the embedding function, and $h(\cdot;\phi)$ represents the hypernetwork. The weights $\theta_{\mathbf{t},\mathbf{y}}$ for each treatment-outcome pair $(\mathbf{t},\mathbf{y})$ are generated by the hypernetwork, rather than learned directly. Consequently, H-Learner is a meta-model approach that naturally accommodates composite treatments and outcomes through dynamic weight-sharing across the treatment-outcome space.

\section{EXPERIMENTS}
\label{sec_experiments}
Here, we present our experimental setup and the corresponding results.

\subsection{Experimental Setup}
\label{subsec_experimental_setup}
\textbf{Baselines:} 
For composite treatments comprising binary treatments only, we compare against DEC, SCP, S-Learner, and VSR. For composite treatments that may be binary, continuous, or categorical, we use S-Learner and xS-Learner, due to the lack of methods which could handle arbitrary treatments. With respect to composite outcomes, we apply these baselines independently for each outcome because none of them are explicitly designed to handle multiple outcomes simultaneously. Specifically, we use a multitasking version of S-Learner (still referred to as S-Learner) and repeat the learner for each outcome referred to as xS-Learner. Unless otherwise specified, our experiments involve five treatments (either all binary or one continuous and four binary) and two outcomes.

\textbf{Metric:} We employ an extension of the Precision in the Estimation of Heterogeneous Effects (PEHE) \cite{hill2011bayesian} suitable for composite treatments and outcomes. Concretely, this extension reduces to computing the mean squared error (MSE) of the estimated treatment effects overall treatment combinations and outcomes.

\textbf{Dataset:} 
Because of the ``fundamental problem of causal inference'' \cite{holland1986statistics}, not all potential outcomes are observable in real-world data, making direct validation of causal methods inherently difficult. As is common in the literature, we use synthetic datasets, where all potential outcomes are artificially generated, to evaluate our proposed methodology. Our data generation process follows \cite{chauhan2023adversarial} and is extended to accommodate multiple treatments and multiple outcomes.

\textbf{Hyperparameters:} 
We implemented SCP, VSR, and DEC using the publicly available code from \cite{qian2021estimating}, adopting their hyperparameter tuning protocols. For H-Learner, we set the learning rate to 0.005, the embedding size to 32, and used a hypernetwork with two hidden layers of 100 neurons each. All other hyperparameters, including the optimiser and batch size, are the same as those used in the baseline implementations.

\subsection{Results}
\label{subsec_results}
In this subsection, we compare H-Learner\footnote{The finalised code will be available at \url{https://github.com/jmdvinodjmd/HLearner}} under two scenarios: (1) arbitrary composite treatments and (2) binary composite treatments.

\textbf{Arbitrary composite treatments:} Since H-Learner shares information across all treatments and outcomes, we compare it against S-Learner \cite{kunzel2019metalearners}, which employs a multitasking neural network, and xS-Learner, which considers only a single outcome at a time and therefore does not share information across outcomes. Figure~\ref{fig_results_continuous} summarises performance as we vary the number of patients, treatments, and outcomes. H-Learner consistently performs better across different data sizes and different numbers of treatments and outcomes. Notably, its advantage over the baselines becomes more pronounced when the dataset size decreases or when the number of outcomes increases. This improvement arises from H-Learner’s dynamic end-to-end information sharing across treatments and outcomes, facilitated by hypernetworks \cite{chauhan2024brief}.

\textbf{Binary composite treatments:}
Here, we restrict treatments to binary treatments and compare H-Learner against S-Learner, VSR, SCP, and DEC. Figure~\ref{fig_results_binary} shows that H-Learner typically outperforms or is on par with the baselines. Its superior performance in smaller datasets, relative to SCP, can be attributed to the way H-Learner shares information across outcomes, which other methods do not. The same mechanism also enables H-Learner to outperform the baselines when the number of outcomes is increased. Although SCP shows stronger results for certain treatments and is overall the second-best estimator, it relies on knowing the underlying causal structure among treatments -- information we provide in these experiments but which may be unavailable in practice. In contrast, H-Learner can accommodate an arbitrary number of treatments (and their interrelationships) by generating a distinct learner network for each treatment combination.

\section{CONCLUSION}
\label{sec_conclusion}
In this paper, we proposed \emph{H-Learner}, a novel hypernetwork-based framework for ITE estimation under composite treatments and composite outcomes -- an important yet underexplored problem in real-world causal inference. By mapping each treatment-outcome combination to a distinct target learner, H-Learner combines the flexibility of independent models with the benefits of a joint learner capable of dynamic end-to-end information sharing. Our empirical results highlight that H-Learner not only manages data scarcity effectively but also consistently outperforms or competes with existing methods when dealing with both arbitrary and binary composite treatments, as well as multiple outcomes. Incorporating hypernetworks enables adaptive weight generation, facilitating transfer learning across complex treatment-outcome configurations and enhancing robustness in scenarios with limited data. We believe H-Learner provides a strong foundation for addressing increasingly intricate real-world causal inference tasks involving multiple treatments and outcomes.



\subsection*{Acknowledgement}
This work was supported in part by the National Institute for Health Research (NIHR) Oxford Biomedical Research Centre (BRC) and in part by InnoHK Project Programme 3.2: Human Intelligence and AI Integration (HIAI) for the Prediction and Intervention of CVDs: Warning System at Hong Kong Centre for Cerebro-cardiovascular Health Engineering (COCHE).
DAC was supported by an NIHR Research Professorship, an RAEng Research Chair, the InnoHK Hong Kong Centre for Cerebro-cardiovascular Health Engineering (COCHE), the NIHR Oxford Biomedical Research Centre (BRC), and the Pandemic Sciences Institute at the University of Oxford. GN is funded by NIHR (Grant number 302607) for a doctoral research fellowship. The views expressed are those of the authors and not necessarily those of the NHS, the NIHR, the Department of Health, the InnoHK – ITC, or the University of Oxford.




\bibliographystyle{plain}
\bibliography{ML4HC}

\end{document}